\documentclass[letterpaper, 10 pt, conference]{ieeeconf}  

\IEEEoverridecommandlockouts                              

\usepackage{graphics} 
\usepackage{times} 
\usepackage{amsmath} 
\usepackage{amssymb}  
\usepackage{comment}
\usepackage{tabularx}
\usepackage{tabularray}
\usepackage{array}
\usepackage{booktabs}
\usepackage{multirow}
\usepackage{graphicx}
\usepackage{hyperref}
\usepackage{subcaption}
\usepackage{xcolor}

\title{\LARGE \bf
Towards Surgical Context Inference and Translation to Gestures
}

\author{Kay Hutchinson$^{1}$$^{*}$,  Zongyu Li$^{1}$$^{*}$, Ian Reyes$^{2}$$^{*}$, Homa Alemzadeh$^{1}$ 
\thanks{* denotes equal contribution.}
\thanks{$^{1}$ Department of Electrical and Computer Engineering, University of Virginia, Charlottesville, VA 22903 USA.
$^{2}$ Ian Reyes was with the Department of Computer Science, University of Virginia, Charlottesville, VA 22903 USA. He is now with IBM.
        {\tt\small \{kch4fk, zl7qw, ir6mp, ha4d\}@virginia.edu} Code for this paper is available at \url{https://github.com/UVA-DSA/Auto_Surgical_Context2Gesture}}%
}

\begin{document}
\newcolumntype{P}[1]{>{\centering\arraybackslash}p{#1}}
\newcolumntype{L}[1]{>{\arraybackslash}p{#1}}
\maketitle
\thispagestyle{empty}
\pagestyle{empty}

\begin{abstract}
Manual labeling of gestures in robot-assisted surgery is labor intensive, prone to errors, and requires expertise or training. We propose a method for automated and explainable generation of gesture transcripts that leverages the abundance of data for image segmentation
. Surgical context is detected using  segmentation masks by examining the distances and intersections between the tools and objects. Next, context labels are translated into gesture transcripts using knowledge-based Finite State Machine (FSM) and data-driven Long Short Term Memory (LSTM) models. We evaluate the performance of each stage of our method by comparing the results with the ground truth segmentation masks, the consensus context labels, and the gesture labels in the JIGSAWS dataset. Our results show that our segmentation models achieve state-of-the-art performance in recognizing needle and thread in Suturing and we can automatically detect important surgical states with high agreement with crowd-sourced labels (e.g., contact between graspers and objects in Suturing). We also find that the FSM models are more robust to poor segmentation and labeling performance than LSTMs. Our proposed method can significantly shorten the gesture labeling process ($\sim$2.8 times). 

\end{abstract}

\section{INTRODUCTION}


Surgical robots for minimally invasive surgery (MIS) 
enable surgeons to operate with greater flexibility and precision, thus reducing incision size, recovery time, and scarring. 
Their widespread adoption into surgical specialties such as urology, gynecology, and general surgery has opened up new fields of interdisciplinary research.
Gesture segmentation and classification has been one of those research areas where both supervised 
\cite{ahmidi2017dataset,lea2016temporal,lea2016segmental,dipietro2016recognizing, dipietro2019segmenting, funke2019using} and unsupervised learning \cite{noy2015unsupervised, fard2016soft,krishnan2017transition, jones2019zero, clopton2017temporal} approaches have been developed for gesture recognition. However, these approaches either rely on black-box deep learning models that are hard to verify and need extensive training data or do not capture the human interpretable contextual information of the gestures.

The JIGSAWS dataset \cite{gao2014jhu} with its surgical gesture labels has been the foundation of many advancements in surgical gesture recognition \cite{van2021gesture}, surgical process modeling \cite{ahmidi2017dataset}, skill assessment \cite{tao2012sparse, varadarajan2009data}, error detection \cite{yasar2020real,Li2022Runtime}, and autonomy \cite{ginesi2021overcoming}. 
However, unlike annotations for surgical instrument segmentation, annotations for surgical workflow such as gestures need guidance from surgeons \cite{Kitaguchi2021Artificial}.
Labeling using descriptive gesture definitions is tedious and subjective, leaving uncertainty as to exactly when gestures start and end, and can have annotation errors that can adversely impact machine learning models and analyses \cite{van2021gesture,hutchinson2021analysis}. 
Recent studies using the JIGSAWS dataset have found errors in $\sim$2-10\% of the gesture labels \cite{van2020multi, hutchinson2021analysis}. 
As emphasized in \cite{van2021gesture}, larger labeled datasets using a common surgical language are needed to support collaboration and comparative analysis.



Some recent works have focused on finer-grained surgical actions such as action triplets \cite{meli2021unsupervised, li2022sirnet, nwoye2022rendezvous} and motion primitives \cite{COMPASS} based on the interactions between robotic tools and objects in the surgical environment. \cite{COMPASS} presented a formal framework for modeling surgical tasks with a unified set of motion primitives that cause changes in surgical context captured from the physical environment. 
These motion primitives were shown to be generalizable across different surgical tasks and can be used to combine data from different datasets. 
\cite{COMPASS} suggests a relation between context and existing gesture labels, but does not define direct relations between the two.


Furthermore, despite limited availability of datasets that include kinematic data from surgical robots, datasets for instrument and object segmentation in MIS procedures are plentiful and have been the subject of imaging competitions \cite{allan20192017,allan20202018}. We propose methods that leverage the abundance of data with image annotations for surgical instruments and important surgical objects to address the challenges of manual labeling and relate surgical context to gestures.
Our goal is to develop an automated, independent, and explainable way of generating gesture transcripts based on video data that does not rely on expensive training data on gestures. Such a method would be easier to verify by humans/experts and can be used as the ground truth for evaluating the black-box gesture recognition models that directly detect gestures from kinematic data.

\begin{figure*}[ht!]   
    \centering
    \begin{subfigure}{0.6\textwidth}
        \centering
        \includegraphics[trim = 0in 3.8in 7.1in 0in, clip, width=\textwidth]{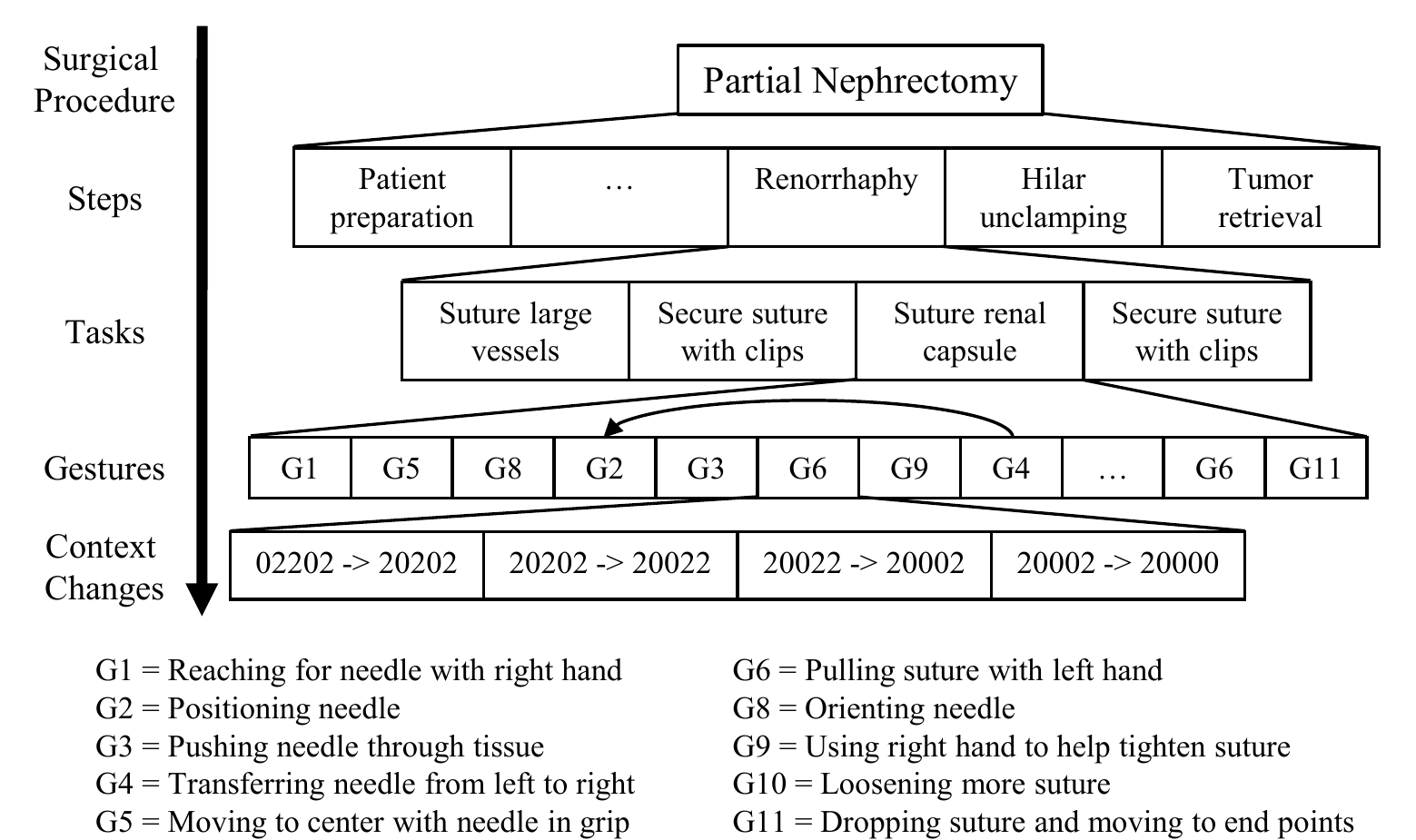}
        \caption{}
        \label{fig:hierarchy}
    \end{subfigure}
    \hfill
    \begin{subfigure}{0.38\textwidth}
        \centering
        \includegraphics[trim = 6.3in 3.8in 2.75in 0in, clip, width=\textwidth]{Figures/hierarchy_context.pdf}
        \caption{}
        \label{fig:context}
    \end{subfigure}
    \vspace{-0.5em}
    \caption{\ref{fig:hierarchy} Surgical hierarchy and relation between gestures and context in a suturing task. \ref{fig:context} State variables and object encodings that comprise context for the JIGSAWS tasks (see Figure \ref{fig:pipeline}). In the Suturing and Needle Passing tasks, a needle is used to throw four sutures through the fabric and rings, respectively, while two knots are tied in the Knot Tying task. 
    }
    \vspace{-1.5em}
    \label{fig:hierarchyandcontext}
\end{figure*}

The main contributions of the paper are as follows:
\begin{itemize}
    \item We present a method for the automated inference of surgical context based on detecting important surgical tool and object interactions using image segmentation.
    \item We propose two methods for automated translation of context labels to gesture labels based on a knowledge-based finite state machine model and a data-driven machine learning model.
    \item We use the JIGSAWS dataset as a case study to demonstrate that our proposed approach results in shorter labeling time using the segmentation masks. 
    
\end{itemize}


\section{PRELIMINARIES}

\subsection{Surgical Process Modeling}

Surgical process modeling \cite{neumuth2011modeling} 
defines how surgical procedures can be decomposed into steps, tasks, and gestures as shown in Figure \ref{fig:hierarchy}. Gestures are defined as actions with semantic meaning for a specific intent and involve particular tools and objects. Thus, they explicitly include the surgical context, capturing important states and interactions in the physical environment. The formal framework in \cite{COMPASS} extended this hierarchy to further include the finer-grained motion primitives (or verbs in action triplets~\cite{nwoye2022rendezvous, neumuth2006acquisition}) as the atomic units of surgical activity (e.g., grasp, push) that lead to changes in context, without explicitly including the semantics of physical context (e.g. needle through tissue). 



\subsection{Surgical Context}
Surgical context is defined as a set of state variables describing the status of a task and interactions among the surgical instruments, objects, and anatomical structures in the physical environment \cite{yasar2019context, yasar2020real, COMPASS}. As shown in Figure \ref{fig:context}, the first four state variables represent objects held by or in contact with the surgical instruments and are the general state variables for all tasks. 
The fifth state variable is task-specific and represents task progress; i.e., the needle's relation to the fabric or ring in the Suturing and Needle Passing tasks, or the knot's status in the Knot Tying task. Figure \ref{fig:context} shows the general and task-specific state variables with their possible values in the Suturing and Knot Tying tasks of the JIGSAWS dataset. 
In Figure \ref{fig:context}, the example context of $ 00202 $ in the Suturing task means that the right grasper is holding the needle and the needle is in the fabric.

The COMPASS dataset \cite{COMPASS} has context labels for all three tasks in the JIGSAWS dataset based on consensus among three annotators. 
But, it does not provide translations from context or motion primitives to gestures which limits comparisons to existing works. Manual labeling was needed to create the context labels which is still subjective and time consuming, despite achieving near-perfect agreement with expert surgeons. 
However, \cite{COMPASS} showed that high quality surgical workflow labels can be generated by examining state variables that comprise the context. With recent improvements in surgical scene segmentation, we show that context can be detected automatically from video data.

\subsection{Surgical Scene Segmentation}

\begin{figure*}[t!]   
    \centering
    \includegraphics[trim = 0in 3.25in 0.5in 0in, clip, width=\textwidth]{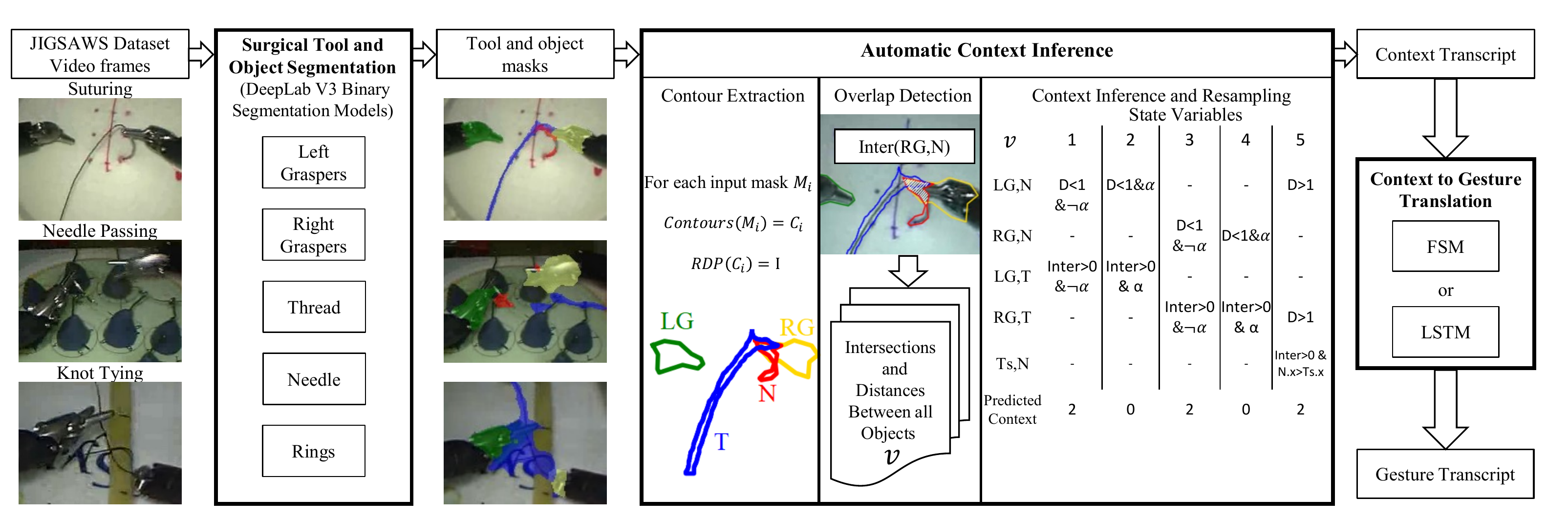}
    \vspace{-1.5em}
    \caption{Pipeline for automatic context inference based on segmentation of video data and context to gesture translation. 
    }
    \vspace{-1.5em}
    \label{fig:pipeline}
\end{figure*}

To advance analysis on video data and provide insights on surgeon performance, the 2017 and 2018 EndoVis workshops at MICCAI introduced a challenge to perform robotic instrument and scene segmentation using images from a da Vinci Xi robot in porcine procedures \cite{allan20192017}. Various models have been proposed in the challenge, but segmenting all objects in a surgical scene has been challenging. 
The DeepLab V3+ model \cite{chen2018encoder} 
achieved the best overall performance in \cite{allan20202018} (see Table \ref{tb:segmentation}). Other DeepLab models \cite{chen2017deeplab, chen2018encoder} have also shown promise in surgical tool and object segmentation.

Most existing works on robot instrument or surgical scene segmentation were based on real surgery videos using publicly available datasets such as MICCAI EndoVis 17 \cite{allan20192017}, MICCAI EndoVis 18 \cite{allan20202018} and Cata7 \cite{ni2019raunet}. Popular frameworks include UNet \cite{ronneberger2015u}, TernausNet \cite{iglovikov2018ternausnet}, and LinkNet \cite{chaurasia2017linknet}. 
Surgical scene segmentation in the dry-lab settings with the JIGSAWS dataset was done in \cite{andersen2021real} and \cite{papp2022surgical}, but we go further by segmenting additional objects and using tool and object segmentation for context inference.
Although surgical scene segmentation and instrument tracking can be used for skill assessment \cite{jin2018tool}, they have not yet been used for automatic context and gesture inference. Hence, our approach could be used as an independent source to evaluate context or gesture segmentation models trained using kinematic data. 

Further, we aim to integrate data-driven segmentation with knowledge-driven context inference and context to gesture translation to perform gesture recognition. Compared to the above deep learning approaches for gesture recognition, this approach enables improvements by integrating human input. Our method also benefits from the availability of large open source image segmentation datasets that provide pretrained weights for segmentation models and could also improve segmentation performance via fine-tuning on smaller datasets.

\section{METHODS}

This section presents our overall pipeline for the automated inference of surgical context and translation to gesture labels based on the video data as depicted in Figure \ref{fig:pipeline}. Surgical context 
can be inferred from the video or kinematic data by estimating the values of the state variables. In this work, we specifically focus on context inference solely based on video data as an independent method to verify gestures predicted from kinematic data or when kinematic data is not available. Our methods are presented for a case study of the JIGSAWS dataset~\cite{JIGSAWS} using the context labels from \cite{COMPASS}, but are applicable to other datasets and sets of gestures. 


\subsection{Tool and Object Segmentation}
The detection of general and task-specific state variables for surgical context requires identifying the status and relative distance of the instruments and the objects of interest in a task. As shown in Figure \ref{fig:context} for the JIGSAWS tasks, these include the left and right graspers, needle, thread, and rings.

We modified the Deeplab V3 model \cite{chen2017deeplab} to perform binary segmentation that classifies the background vs. one object class in the video frames of a task trial. Specifically, we train separate binary classification models to classify background vs. left grasper, background vs. right grasper, background vs. needle, background vs. thread, and background vs. ring. The input to each model is a matrix $A_{H \times W \times 3}$  representing an RGB image of a video frame with Height (H) and Width (W). The output is a binary matrix $M_{H \times W}$ representing the segmentation mask with 0 for the background class, and 1 for the segmented object class.  We need to infer the intersections between objects for generating context, which cannot be done with the existing multi-class segmentation models that classify each pixel to a single object class.
Binary segmentation models for each object class enable the analysis of intersections and overlaps among separate object masks to infer interactions between objects.

For each object, we combine the data from all tasks to train a single model to classify that object in all tasks. We leveraged transfer learning by initializing the model with a ResNet-50 \cite{he2016deep} backbone pre-trained on the COCO dataset \cite{lin2014microsoft}. We obtained tool and object annotations for the JIGSAWS dataset and used a subset of 70 videos for fine-tuning the model. However, the test set for the whole pipeline was significantly limited since much of the data from JIGSAWS was needed to train the image segmentation models. We trained our models for up to 20 epochs using Adam optimization \cite{kingma2014adam} with a learning rate of $10^{-5}$. 



\subsection{Automated Context Inference}
The masks from the segmentation models provide us with information about the area and position of the instruments and objects 
which can 
enable 
state variable estimation  
at each frame. By calculating intersections and distances between the object masks in a given frame, we can detect interactions such as \textit{contact} and \textit{hold} as shown in Figure \ref{fig:context}. 

In the mask matrices $M_{H \times W}$ generated by the segmentation models, each element $m_{hw} \in \{0,1\}$ indicates if the pixel $(h,w)$ belongs to an object mask. We first perform a pre-processing step on $M$ to eliminate the noise around masks such as the needles and threads. 
Contour extraction is done to help eliminate the rough edges of the masks and improve intersection detection. This step uses the OpenCV library \cite{opencv_library} to iteratively construct contours around every element $m_{hw} \in M$, thus reducing the input matrix to a list of points $p \in C \subset M $ for each instrument class where $C$ is the boundary of $M$. 
Using simplified polygons instead of binary masks greatly reduces the time needed to calculate intersections and distances between objects for each frame.
We experimentally determined that dropping polygons with areas under 15 pixel units squared and smoothing the polygons using the Ramer–Douglas–Peucker (RDP) algorithm \cite{RAMER1972244,douglas1973algorithms} results in better accuracy based on training set.

Next, we detect overlaps between masks by taking a list of valid polygons and calculating a feature vector $v$ of distances ($D$) and intersection areas ($Inter$) between pairs of input masks. The input polygons Left Grasper $(LG)$, Right Grasper $(RG)$, Thread $(T)$ are common for all tasks. 
Task-specific objects are the Needle $(N)$ appearing in Needle Passing and Suturing, the manually labeled Tissue Points $(Ts)$ representing the markings on the tissue where the needle makes contact in Suturing, and the Rings $R$ in Needle Passing. 

We define the distance functions $D(I, J)$ and $d(i,j)$ and the intersection function $Inter(I,J)$ to, respectively, calculate the pixel distance between two object masks $I$ and $J$, the pixel distance between the individual polygons $i_1, j_1, ...$ that constitute an object mask, and the area of intersection between two object masks $I$ and $J$. For any object polygon $I$ which is comprised of several polygon segments $i_1, i_2, ..., i_n$, the distance to any other object $J$ can be calculated as: $D(I, J) = \text{average}([d (i,j) \text{ for } i \in  I \text{ and }  j \in  J])$. The intersection function $Inter(I,J)$ is implemented using a geometric intersection algorithm from the Shapely \cite{shapely} library. We also define the components $I.x,I.y$ for an object I as the horizontal and vertical coordinates of the midpoint of its polygon $I$, calculated as the average of every point in $I$.  
In order to determine the Boolean function $ (\alpha) $ for each grasper, if the distance between the manually labeled pixel coordinates of the grasper jaw ends was less than 18 pixels, then the grasper was closed ($ \neg \alpha $), else it was open ($ \alpha $). 


\setlength{\abovedisplayskip}{-3pt}
\setlength{\belowdisplayskip}{-3pt}
\small
\begin{align} 
\text{Left Hold} &
\begin{cases} \label{equn:LH}
    2   &   \text{if } D(LG,N)<1 \wedge \neg \alpha \\ 
    3   &   \text{if } Inter(LG,T)>0 \wedge \neg \alpha \\
    0   &   \text{otherwise}
\end{cases}
\\
\text{Left Contact} &
\begin{cases} \label{equn:LC}
    2   &   \text{if } D(LG,N)<1 \wedge \alpha \\
    3   &   \text{if } Inter(LG,T)>0 \wedge \alpha \\
    0   &   \text{otherwise} 
\end{cases}
\\
\text{Needle} &
\begin{cases} \label{equn:N}
    2   &   \text{if} (Inter(Ts,N) > 0 \wedge N.x < Ts.x)  \\
    1   &   \text{if} (Inter(Ts,N) = 0 \vee N.x\geq Ts.x) \wedge  \\
        &   (D(RG,T)>1 \vee D(LG,N)>1) \\  
    0   &   \text{otherwise} 
\end{cases}
\end{align}
\normalsize
\setlength{\abovedisplayskip}{6pt}
\setlength{\belowdisplayskip}{6pt}

 The feature vector $v=< D(LG,N),Inter(LG,T),...>$ (see Figure \ref{fig:pipeline}) is then used to estimate the values of different state variables using a set of task-specific functions. 
 An example set of functions is shown in Equations \ref{equn:LH}-\ref{equn:N} for the state variables relating to the left robot arm and needle in Suturing task. A similar set of functions are used for the right arm. For example, if the distance between the left grasper and needle is less than one pixel ($D(LG,N)<1$) and the grasper is closed ($\neg \alpha$), then a value of 2 is estimated for the \textit{Left Hold} variable. Or the \textit{Needle} state is detected as touching (2) when the relative horizontal distance of the needle polygon $(N.x)$ is less than the average (midpoint) of the tissue points $(Ts.x)$ and these two objects intersect ($Inter > 0$).  
The input sample rate of the context to gesture translation was 3Hz, so the final estimated variables were downsampled from 30Hz to 3Hz using a rolling mode for each state variable with a window of 10 frames.

\subsection{Context to Gesture Translation}
The last step in our pipeline translates the automatically generated context labels into gesture labels. 
The input to the translation model is a 2-dimensional time series matrix $\chi_{State \times n}$, where $State$ represents the 5 state variables describing the context (see Figure \ref{fig:context}) and $n$ represents the total number of samples in the trial. %
We map each time step $State_t$ to a corresponding gesture $G_i$ in the JIGSAWS dataset. The translation output is a 1-dimensional time series $Y_{n} \in \{\mathbb{G}\}$ with each time step mapped to a gesture.
We present two approaches based on domain knowledge and data. 

\subsubsection{Finite State Machine Model}
\begin{figure}[b!]
    \vspace{-1.75em}
    \centering
    \includegraphics[trim = 0in 5.45in 8.85in 0in, clip, width=0.49\textwidth]{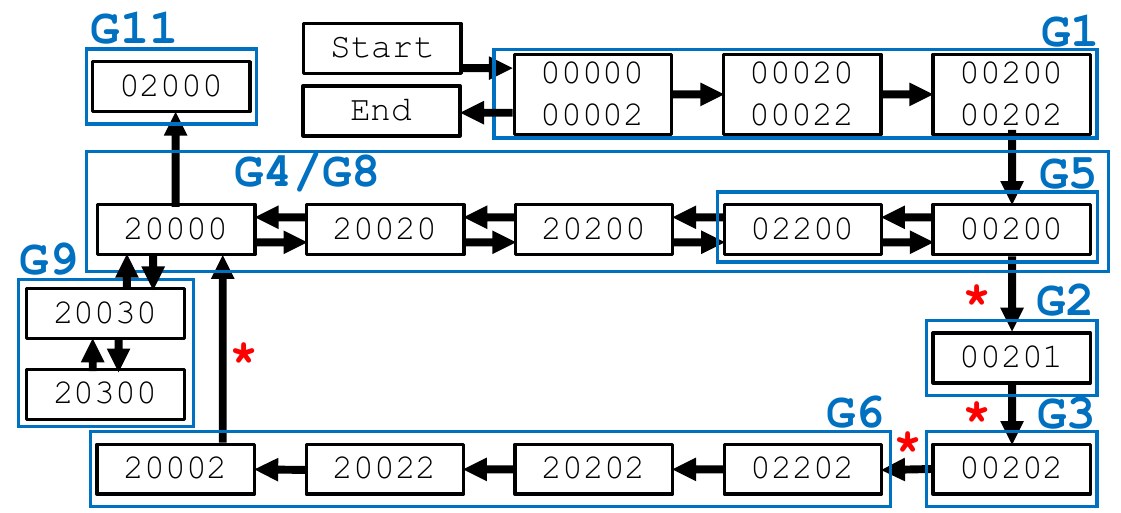}
    \caption{Grouping and mapping of context to gestures in the grammar graph of the Suturing task. \textcolor{red}{*} denotes transitions due to duration limits as follows: G2$>$6.0 s $\rightarrow$ G3, G3$>$11.1 s $\rightarrow$ G6, G4$>$5.2 s $\rightarrow$ G2, G6$>$6.1 s $\rightarrow$ G4.}
    \label{fig:contextgestures}
    \vspace{-0.25em}
\end{figure}
Our first approach relies on a finite state machine (FSM) defined based on the knowledge of surgical tasks which directly relates context to gestures and is more explainable than deep learning models. The grammar graphs from \cite{ahmidi2017dataset} for each task were overlaid on top of the ideal context models from \cite{COMPASS} so that each gesture could be mapped into the groups of contextual changes that happen as the result of executing the gesture (see Figure \ref{fig:contextgestures} for the Suturing task). For example, G2 (positioning needle) corresponds to a change from a `0' to a `1' in the fifth state variable. 
Or G4 (transferring needle from left to right) is the context sequence $ 20000 \rightarrow 20020 \rightarrow 20200 \rightarrow 02200 \rightarrow 00200 $ which means the needle is initially held in the left grasper, then touched and grasped by the right grasper, and released by the left grasper. In Figure \ref{fig:contextgestures}, the G4 and G8 groupings overlap since G8 (orienting needle) is performed by passing the needle from the right to the left grasper and back to the right grasper while changing its orientation. 

Given the context transcript of a trial, the FSM is evaluated for each context and a transition to the next gesture is detected if the input context is part of the next gesture. 
The FSM for each task was initialized in the `Start' state since not all of the trials started with G1. Also, G11 was assumed to be last and so it was appended to the gestures following the last detected gesture. 
In addition, in the Suturing and Needle Passing tasks, G9 and G10 had low rates of occurrence and were not included in the final translation. This allowed us to focus only on state changes involving the needle and thus ignore grasps and touches of the thread and rings with the added benefit of simplifying the FSMs and limiting the total number of valid context changes.

We also consider gesture duration as a trigger for transitions between gestures. If the current gesture's duration exceeds a certain threshold based on the average 
duration of that gesture class, a transition to the next gesture is enforced. This is to address the cases where a gesture transition does not happen due to inaccuracies in context detection. For example, the segmentation models tend to have lower accuracy in detecting the needle and thread states, leading to not detecting transitions that are dependent on those states. 

\subsubsection{LSTM Model}
Our second approach for translation of context to gesture transcripts relies on sequential deep learning methods to learn relationships in the data that are not captured by the FSM models. We trained an LSTM model to perform automated context to gesture translation for each task. We chose the LSTM model for its ability to learn temporal features. Specifically, we used a simple double layer LSTM network with 64  hidden units for the Suturing and Needle Passing tasks and 256 hidden units for the Knot Tying task. We used Adam optimization \cite{kingma2014adam} and the cross entropy loss function to train the models. The hidden layers, number of hidden units and learning rates were determined by hyperparameter tuning. The final models were trained with the best model configurations and used to perform inference on the automatically generated context labels using the segmentation masks in the test set. Note that the LSTM model is a black box model and does not provide transparency like the FSM model in the previous section. 



\section{EXPERIMENTAL EVALUATION}
\subsection{Experimental Setup}
We use an 80/20 train/test split of the JIGSAWS dataset for evaluating our pipeline. The original videos are 30Hz and we obtained binary masks for the tools and objects at 2Hz which we then used to train/test the segmentation models.
The LSTM networks are trained with the 3Hz context labels from \cite{COMPASS}. 
We evaluate both the FSM and LSTM for context to gesture translation with the test set context labels.




The experiments were conducted on a PC with an Intel Core i7 CPU \@ 3.60GHz, 32GB RAM, and an NVIDIA GeForce RTX 2080 Ti GPU running Ubuntu 18.04.2 LTS.

\subsection{Metrics}
The following metrics were used to evaluate the pipeline. 

\textbf{Accuracy:}
Accuracy is the ratio of samples with correct labels divided by the total number of samples in a trial.

\textbf{Edit Score:}
Edit score is calculated using Equation \ref{equn:edit} from \cite{lea2016temporal} where the normalized Levenshtein edit distance, $ edit(G, P) $, quantifies the number of insertions, deletions, and replacements needed to transform the sequence of predicted labels $ P $  to match the ground truth sequence of labels $ G $. This is then normalized by the maximum length of the sequences so that a higher edit score is better. 

\small
\begin{equation}
    \text{Edit Score} = (1-\frac{edit(G, P)}{max(len(G), len(P))}) \times 100
    \label{equn:edit}
\end{equation}
\normalsize

\textbf{Intersection over Union (IOU):}
Mean IOU, as calculated in Equation \ref{equn:IOU}, is the standard for assessing the segmentation and translation models \cite{lin2014microsoft}. 

\small
\begin{equation}
    IOU = TP/(TP+FP+FN)
    \label{equn:IOU}
\end{equation}
\normalsize

Each predicted segment is matched to a corresponding segment in the ground truth. Then, the average IOU for each class is calculated and the mean of class IOUs is returned.


\subsection{Results}
\begin{table}[t!]
\centering
\caption{Tool and object segmentation performance on the test set (mean IOU for each object class) on the MICCAI18 (M) and JIGSAWS Suturing (S), Needle Passing (NP), and Knot Tying (KT) tasks.}
\vspace{-0.5em}
\label{tb:segmentation}
\begin{tabular}{P{0.275\linewidth} P{0.06\linewidth} P{0.06\linewidth} P{0.06\linewidth} P{0.075\linewidth} P{0.075\linewidth} P{0.05\linewidth}}
\toprule
\multirow{2}{*}{Model} & \multirow{2}{*}{Data} & \multicolumn{ 2}{c}{Graspers} & \multicolumn{ 3}{c}{Objects} \\
 & & Left & Right & Needle & Thread & Ring \\
\midrule 
Deeplab V3+ \cite{allan20202018} & \multirow{2}{*}{M~\cite{allan20202018}} & \multicolumn{ 2}{c}{0.78} & 0.014 & \textbf{0.48} & N/A \\
U-net \cite{allan20202018} & & \multicolumn{ 2}{c}{0.72} & 0.02 & 0.33 & N/A \\ \midrule
Mobile-U-Net~\cite{andersen2021real} & S & \multicolumn{ 2}{c}{\textbf{0.82}} & \textbf{0.56} & N/A & N/A \\ \midrule
\multirow{2}{*}{Trained UNet \cite{papp2022surgical}} & S &\multicolumn{ 2}{c}{0.69} & N/A & N/A & N/A \\
 & NP &\multicolumn{ 2}{c}{0.66} & N/A & N/A & N/A \\
Trained LinkNet34  & KT & \multicolumn{ 2}{c}{0.80} & N/A & N/A & N/A \\ \midrule
\multirow{3}{*}{Deeplab V3 (ours)} & S & 0.71 & \textbf{0.64} & \textbf{0.19} & \textbf{0.52} & N/A \\ 
 & NP & 0.61 & 0.49 & 0.09 & 0.25	 & \textbf{0.37} \\ 
 & KT & \textbf{0.74} & 0.61 & N/A & 0.44 & N/A \\ 
\bottomrule
\end{tabular}
\vspace{-2.5em}
\end{table}

\subsubsection{Tool and Object Segmentation}
\label{sec:res tool and obj seg}

\begin{table*}[ht!]
\centering
\caption{State variable IOU with consensus context using predicted masks from DeepLab V3 and ground truth masks}
\vspace{-0.5em}
\label{tab:generation}
\begin{tabular}{P{1.8cm}  P{0.7cm} P{1cm} P{0.7cm} P{1cm} P{1cm} | P{0.7cm}  P{0.7cm} P{1cm} P{0.7cm} P{1cm} P{1cm} | P{0.7cm} }
 & \multicolumn{6}{c}{Predicted Masks} & \multicolumn{6}{c}{Ground Truth Masks} \\
\cmidrule(lr){2-7} \cmidrule(lr){8-13}
 & Left Hold & Left Contact & Right Hold & Right Contact & Needle or Knot & Avg & Left Hold & Left Contact & Right Hold & Right Contact & Needle or Knot & Avg \\
\cmidrule(lr){1-7} \cmidrule(lr){8-13}
Suturing & 0.48 & 0.75 & \textbf{0.60} & 0.87 & 0.30 & 0.60 & 0.52 & 0.77 &  \textbf{0.61} &  0.87 &  0.39 & 0.63\\ 
Needle Passing & 0.40 & \textbf{0.97} & 0.18 & \textbf{0.95} & \textbf{0.39}& 0.58 & 0.42 &  \textbf{0.97} &  0.19 & \textbf{0.94} & \textbf{0.41} & 0.59\\
Knot Tying & \textbf{0.75} & 0.72 & 0.57 & 0.78 & \textbf{0.59} & \textbf{0.68} & \textbf{0.83} & 0.77 & \textbf{0.61} & 0.79 & \textbf{0.62} & \textbf{0.72}\\
\cmidrule(lr){1-7} \cmidrule(lr){8-13}
Avg & 0.54 & \textbf{0.81} & 0.45 & \textbf{0.87} & 0.43 & & 0.59 & \textbf{0.84} & 0.47 & \textbf{0.87} & 0.47 & \\ 
\cmidrule(lr){1-7} \cmidrule(lr){8-13}
\end{tabular}
\end{table*}

\begin{table*}[h!]
\centering
\caption{Gestures translated from the automatic context inference given masks from the Deeplab V3 models } 
\vspace{-0.5em}
\label{tab:c2g_all}
\begin{tabular}{P{1.75cm} P{1.75cm} P{1.65cm} P{1.65cm}  P{1.65cm} P{1.65cm} P{1.65cm} P{0.65cm}}
 & & \multicolumn{3}{c}{Gestures from Predicted Context Labels} & \multicolumn{3}{c}{Gestures from Consensus Context Labels} \\
\cmidrule(lr){3-5} \cmidrule(l){6-8}
Task & Model & Accuracy (\%) & Edit Score & IOU & Accuracy (\%) & Edit Score & IOU  \\ \midrule
\multirow{2}{*}{Suturing} & FSM & \textbf{40.8} & \textbf{67.1} & \textbf{0.28} & \textbf{66.3} & \textbf{84.4} & \textbf{0.48} \\ 
  & LSTM  & 25.4 & 24.9  & 0.17 &  38.8 &  34.7 & 0.26\\ 
\cmidrule(lr){1-2} \cmidrule(lr){3-5} \cmidrule(l){6-8}
Suturing (vid) & Zero-shot \cite{jones2019zero} & \textbf{56.6} & \textbf{61.7} & & & & \\
Suturing (kin) & TSSC-DL \cite{clopton2017temporal} & 49.7 & 32.8 & & & & \\

\cmidrule(lr){1-2} \cmidrule(lr){3-5} \cmidrule(l){6-8}
\multirow{2}{*}{Needle Passing} & FSM & \textbf{18.0}  & \textbf{76.2}  & \textbf{0.12} &  \textbf{70.1} &  \textbf{88.7}  & \textbf{0.54}\\ 
   & LSTM & 14.8 & 20.1 & 0.02 &  17.0 & 20.0 &  0.04\\ 
\cmidrule(lr){1-2} \cmidrule(lr){3-5} \cmidrule(l){6-8}
\multirow{2}{*}{Knot Tying} & FSM & \textbf{42.9} & \textbf{70.7} & \textbf{0.43} &  \textbf{54.6}  &  \textbf{91.5} &  \textbf{0.43}\\ 
   & LSTM & 36.5 & 23.8 & 0.17 & 50.8 &  49.2 &  0.28 \\ 
\bottomrule
\end{tabular}%
\vspace{-1.5em}
\end{table*}

 Table \ref{tb:segmentation} shows the performance of our segmentation models in comparison to the related work. 
 Although the MICCAI 18 challenge \cite{allan20202018} dataset is from real porcine procedures, and differs from the JIGSAWS dataset collected from dry-lab experiments, it has similar objects including the clasper (similar to the graspers in JIGSAWS), needle and thread. 
 The Deeplab V3+ model achieved the best performance on the thread class. The top models from MICCAI 18 do not perform as well as our binary models on the needle and thread classes in the Suturing task. 
 However, the Mobile-U-Net \cite{andersen2021real} achieved the highest performance for grasper and needle segmentation in the JIGSAWS Suturing task. 
 \cite{papp2022surgical} reported tool segmentation IOUs for all the JIGSAWS tasks with up to 0.8 for KT using a Trained LinkNet34, but did not do object segmentation. 
 Among the JIGSAWS tasks, we achieved the best performance in Suturing for the right grasper, needle and thread, while the model performance on the Needle Passing task was the worst. This is likely due to Needle Passing's background having less contrast with the foreground compared to the other two tasks as shown in Figure \ref{fig:pipeline}). We can also see that the needle and thread masks are thinner compared to the grasper masks. So, the mask boundary errors could contribute to a lower score for the needle and thread classes. 
 The estimated time for segmenting the whole JIGSAWS dataset is 8.6 hours.

\subsubsection{Automated Context Inference}
\label{sec:res Automated Context Labeling}
Table \ref{tab:generation} shows the performance of the context inference method in terms of IOU achieved for each state variable with the predicted segmentation masks and the ground truth masks from crowd-sourcing.

The left column of Table \ref{tab:generation} shows that left and right contact have higher IOUs compared to left and right hold, and the needle or knot state has the lowest IOU. 
This is because errors in estimating the position of the grasper jaw ends affect accurate inference of the hold state, while contact is relatively simple by finding if the two masks intersect. Better performance in detecting contact compared to hold states is also observed in the right column of Table \ref{tab:generation}, where ground truth segmentation masks are used. 
Hence, the lower performance of the left hold and right hold could primarily be due to the difficulty in detecting these states.

For the needle/knot state, we need to detect if the needle is in the fabric/tissue for the Suturing task, in/out of the ring for the Needle Passing task, and if the knot is loose or tight in the Knot Tying task. Detecting the state of the needle and knot is difficult even with the ground truth segmentation masks in the right column of Table \ref{tab:generation}. This is because the needle and thread have the lowest segmentation performance compared to graspers as shown in Table \ref{tb:segmentation}.
The total time to perform automatic context inference is estimated to be about 30 seconds for the whole JIGSAWS dataset.

\subsubsection{Context to Gesture Translation}
\label{sec:res Context to Gesture Translation}
The right column of Table \ref{tab:c2g_all} shows the performance of the FSM and LSTM methods in translating ground truth context labels to gestures. 
The FSM model achieves higher accuracies and edit scores than the LSTM. 
The left column of Table \ref{tab:c2g_all} shows the performance of the overall pipeline with automated context labels. We see that using automated context from predicted masks degrades the performance of both models because the segmentation models perform poorly at generating masks for the needle and for all tools and objects in Needle Passing. This effect is propagated through the pipeline, resulting in low accuracies and IOUs. 
The FSM generally outperforms the LSTM likely due to its knowledge-based structure and setting limits on gesture durations that prevent the model from becoming stuck in any one gesture even with degraded context labels.  
The FSM pipeline achieves accuracies lower than unsupervised models from \cite{jones2019zero} and \cite{clopton2017temporal} for Suturing, but outperforms them in terms of edit score. 
These observations suggest that there are benefits to incorporating knowledge into context to gesture translation that can make the model more robust to degraded context labels. 
However, the FSM is manually developed based on domain knowledge and relies on defined inputs and transitions while the LSTM requires labeled data for training. The time to generate the entire JIGSAWS gesture translation from context is less than 3 minutes for both models.

\section{DISCUSSION AND CONCLUSIONS}
Our proposed pipeline for automated inference of surgical context and translation to gesture labels can perform automatic and explainable gesture inference given video segmentation masks. 
It can be used as an efficient and fast inference method by significantly shortening manual gesture labeling time ($\sim$9 hours vs. $\sim$26 hours for the case study of the JIGSAWS dataset). 
We rely on models pre-trained on general images and publicly-available datasets which lowers the 
cost of manually labeling video data and makes our model generalizable to other datasets and tasks. 

For the case study of JIGSAWS, our binary segmentation models achieve comparable performance to state-of-the-art models on the grasper and thread classes, and better performance on the needle class. However, they do not perform well enough for the needle and thread classes which are important for accurate context inference. 
Our context inference method also does not perform equally well for all the states. Given the ground truth segmentation masks, it achieves $\sim$85\% IOU for states such as left/right contact, but only $\sim$45\% IOU for the needle/knot state.
The FSM and LSTM models for context to gesture translation have better performance given ground truth context labels compared to predicted context which may be due to imperfect models at each stage of the pipeline and error propagation. 

Manual annotations for the grasper end points and tissue points were used for context inference. 
Also, our method relies on 2D images to infer context from a 3D environment which can particularly complicate detecting the contact states. 
Future work will focus on addressing these limitations and improving the performance and robustness of the overall pipeline to apply it to runtime error detection \cite{yasar2020real, Li2022Runtime}.

\section*{ACKNOWLEDGMENT}
This work was supported in part by the National Science Foundation grants DGE-1829004 and CNS-2146295.


\bibliographystyle{IEEEtran}
\bibliography{icrabib.bib}

\end{document}